\documentclass[graybox]{svmult}

\usepackage{type1cm}        % activate if the above 3 fonts are
\usepackage{makeidx}         % allows index generation
\usepackage{graphicx}        % standard LaTeX graphics tool
\graphicspath{ {./images/} }
\usepackage{subcaption}
\usepackage{tablefootnote}
\usepackage{xcolor}

\usepackage{multicol}        % used for the two-column index
\usepackage[bottom]{footmisc}% places footnotes at page bottom

\usepackage{newtxtext}       % 
\usepackage{newtxmath}       % selects Times Roman as basic font

\usepackage[numbers]{natbib}

\makeindex             % used for the subject index
\begin{document}

% \title*{Personalized Value Alignment and its Effects on Trust and Team Performance in Human-Robot Teams}， OR 
\title*{Value Alignment and Trust in Human-Robot Interaction: Insights from Simulation and User Study}
% \titlerunning{Personalized Value Alignment in Human-Robot Teams}
\titlerunning{Value Alignment and Trust in Human-Robot Interaction}

\author{Shreyas Bhat, Joseph B. Lyons, Cong Shi, X. Jessie Yang}
\authorrunning{Bhat et al.}

\institute{Shreyas Bhat \at University of Michigan, Ann Arbor, MI, \email{shreyasb@umich.edu}
\and Joseph B. Lyons \at Air Force Research Laboratory, Dayton, OH, \email{joseph.lyons.6@us.af.mil}
\and Cong Shi \at University of Miami, Miami, FL,
\email{congshi@bus.miami.edu}
\and X. Jessie Yang \at University of Michigan, Ann Arbor, MI,
\email{xijyang@umich.edu}}
\maketitle

\abstract*{With the advent of AI technologies, humans and robots are increasingly teaming up to perform collaborative tasks. To enable smooth and effective collaboration, the topic of value alignment (operationalized herein as the degree of dynamic goal alignment within a task) between the robot and the human is gaining increasing research attention. Prior literature on value alignment 
% in human-robot teaming 
makes an inherent assumption that aligning the values of the robot with that of the human benefits the team. This assumption, however, has not been empirically verified. Moreover, prior literature does not account for human's trust in the robot when analyzing human-robot value alignment. Thus, a research gap needs to be bridged by answering two questions: How does alignment of values affect trust? Is it always beneficial to align the robot's values with that of the human? We present a simulation study and a human-subject study to answer these questions. Results from the simulation study show that alignment of values is important for trust when the overall risk level of the task is high. 
% In this case, the misalignment of values between the human and the robot leads to a low level of trust from the human. 
We also present an adaptive strategy for the robot that uses Inverse Reinforcement Learning (IRL) to match the values of the robot with those of the human during interaction. Our simulations suggest that such an adaptive strategy is able to maintain trust \textcolor{blue}{across the full spectrum of human values.}
% regardless of the values of the human. 
% In order to validate these results from simulations, we conducted an empirical study. 
We also present results from an empirical study that validate these findings from simulation. Results indicate that real-time personalized value alignment is beneficial to trust and perceived performance by the human when the robot does not have a good prior on the human's values.
% On the other hand, in the presence of an accurate prior on the human's values, an adaptive interaction strategy focusing on personalized value alignment may have less value compared to a non-adaptive strategy based on said prior.
}

\abstract{With the advent of AI technologies, humans and robots are increasingly teaming up to perform collaborative tasks. To enable smooth and effective collaboration, the topic of value alignment (operationalized herein as the degree of dynamic goal alignment within a task) between the robot and the human is gaining increasing research attention. Prior literature on value alignment 
% in human-robot teaming 
makes an inherent assumption that aligning the values of the robot with that of the human benefits the team. This assumption, however, has not been empirically verified. Moreover, prior literature does not account for human's trust in the robot when analyzing human-robot value alignment. Thus, a research gap needs to be bridged by answering two questions: How does alignment of values affect trust? Is it always beneficial to align the robot's values with that of the human? We present a simulation study and a human-subject study to answer these questions. Results from the simulation study show that alignment of values is important for trust when the overall risk level of the task is high. 
% In this case, the misalignment of values between the human and the robot leads to a low level of trust from the human. 
We also present an adaptive strategy for the robot that uses Inverse Reinforcement Learning (IRL) to match the values of the robot with those of the human during interaction. Our simulations suggest that such an adaptive strategy is able to maintain trust across the full spectrum of human values.
% regardless of the values of the human. 
% In order to validate these results from simulations, we conducted an empirical study. 
We also present results from an empirical study that validate these findings from simulation. Results indicate that real-time personalized value alignment is beneficial to trust and perceived performance by the human when the robot does not have a good \textit{prior} on the human's values.
% On the other hand, in the presence of an accurate prior on the human's values, an adaptive interaction strategy focusing on personalized value alignment may have less value compared to a non-adaptive strategy based on said prior.
\keywords{Human-Robot Teaming, Value Alignment, Trust in HRI}
}

\section{Introduction}
\label{sec:intro}
Robots are progressively becoming more essential in our everyday activities, establishing their roles in numerous fields such as healthcare, manufacturing, education, and domestic aid, among others. With this ongoing integration, robots are transitioning from being viewed merely as devices executing specific tasks to being recognized as cooperative allies that work alongside humans. Within the context of human-robot teaming, research studying the trust of the human on their robot counterpart is becoming increasingly important \cite{sheridan_humanrobot_2016, Billings2012, Chiou2021, Yang:2017:EEU:2909824.3020230, azevedo-sa_unified_2021}. Without an adequate level of trust, we cannot realize the full potential of human-robot teams. 

Substantial research efforts have been directed towards creating robots that demonstrate trustworthy behavior and exploring ways to predict and influence a human's trust in robots. For example, one particular line of inquiry is concentrated on generating explanations for robotic actions \cite{Wang2016, Wang:0tu, Lyons2023, lyons_explanations_2023, DU2019428}, which often results in increased perception of reliability and, consequently, greater trust in these machines. Other research areas are focused on developing algorithms that can predict trust in real-time \cite{Xu2015, Guo2021, guo_enabling_2023, Soh2020, yang_toward_2023}, understanding the dynamics of trust \cite{yang_toward_2023, Cohen_dynamics_2021,  Yang2021_chapter}
% , chung2024associations}, 
devising strategies to mend broken trust \cite{karli_what_2023, esterwood_three_2023, guo_reward_2023}, and creating planning methods that take trust into consideration \cite{Bhat2022, Chen2020, Kumar2019b, Chen2018, zahedi2023, Pippin2014}.

More recently, the idea of value/goal alignment -- aligning the values/goals of robots with those of humans, has garnered significant attention, with the assumption that such alignment would benefit human-robot interaction \cite{sanneman_validating_2023, Yuan2022}. Recent literature in value/goal alignment is primarily focused on enabling the autonomous or robotic agent to learn the human's values/goals through preferences \cite{Hadfield-Menell2016, biyik2018batch, christiano2023deep} or demonstrations \cite{Fisac2020, Hadfield-Menell2016, arora2021}.

However, there is a lack of research empirically examining and quantifying the effects of alignment on human-robot interaction processes and outcomes. Yet, there are at least three reasons to suggest that such alignment could be beneficial. First, prior research has illustrated that agent adaptation to humans can enhance performance  \cite{Azevedo-Sa_2020, luo_workload_2021}.  Second, agent adaptation could be viewed as the agent being responsive to the human and may, in turn, increase human trust in the agent and enhance team performance \cite{Li2021}. Third, value alignment not only could facilitate trust establishment and enhance team performance, but it is also important for ensuring that machine partners are morally acceptable as they negotiate complex social situations \cite{Laakasuo2023}. 

In this chapter, we present results from two studies -- a simulation and an empirical verification of the results from simulation. The goal of these studies was to examine the effect of the degree of alignment of values (seen through the lens of reward weights) between the human and the robot in a task with conflicting goals. The results of the simulation study indicate that misalignment of values between the human and the robot result in a loss of trust, but this loss is only apparent under high risk scenarios. Under low risk scenarios, the degree of misalignment is not very influential on trust. In high risk scenarios, therefore, a robot that adapts its values to that of the human will result in higher trust by the human. We empirically verify this claim through a human-subjects study conducted in a high-risk scenario. The results of this study show that the adaptive-learner strategy presented in our work is able to learn the human's values during interaction and this, build and maintain trust, when compared to a non-adaptive strategy. 

\section{Related Work}
\label{sec:related-work}
There are two bodies of work closely related to the studies presented in this chapter. One deals with using trust as a decision-variable to drive the decision-making processes of robots in human-robot teams and the other deals with ensuring the alignment of values between the human and the robot. A brief review of these fields is presented in this section. 

\subsection{Trust-Aware Decision-Making}
\label{subsec:trust-aware-decision-making}

In the past few years, significant work has been done to create algorithms for robots that take into account the level of trust humans have in them while performing collaborative tasks \cite{washburn_trust-aware_2021, li_trust-aware_2024, hussein_towards_nodate, yu_trust-aware_2023, pang_trust-aware_2021, Guo2021b, Bhat2022}. These studies incorporate a representation of a human's trust in the robot into the decision-making processes of the robot and utilize quantitative models of trust and predict human behavior depending on this level of trust.

A quantitative trust model based on the beta distribution is presented in Guo et al. \cite{Guo2021}. This model was then used in a simulation study done by Guo et al. \cite{Guo2021b} to demonstrate the need for using a trust-gaining reward term in the reward function of the robot to encourage trust-gaining behavior. Bhat et al. \cite{Bhat2022} then conducted a human-subjects experiment using this model and found three types of trust dynamics and associated personal characteristics with the type of trust dynamics exhibited by a person, highlighting the need for robot adaptation given individual differences in trust dynamics.
% Such associations have also been observed in episodic human-robot collaborative tasks \cite{chung2024associations}.

Trust-aware decision making has also been made possible by modeling the interaction as a Partially Observable Markov Decision Process (POMDP) with trust as the partially observable state variable \cite{Kumar2019a, Kumar2019b, Chen2020, yu_trust-aware_2023}. Akash et al. \cite{Kumar2019a} present a trust-workload POMDP model that can be learned through interaction data and can be solved to generate optimal policies for the robot that control the level of transparency of the robot's interface \cite{Kumar2019b}. Chen et al. \cite{Chen2020} provide a trust-POMDP model that can be solved for getting optimal policies for a robot. In a collaborative pick and place task with a robotic arm, they use their model to show trust-gaining behaviors learnt by the robot when it senses that the human's trust is low. On the other hand, when trust is high, the robot selects risky but high-reward actions to gain the maximum reward. 

Zahedi et al. \cite{zahedi2023} formulate a meta-MDP problem with trust as a decision variable to choose the robot's behavior enabling trust-aware decision making for the robot. A robot using this framework is shown to display sub-optimal but trustworthy behaviors over optimal but untrustworthy behaviors when the human's trust is low. At the same time, if there is enough trust, the robot chooses the optimal action plans to maximize reward gain. 

Our work differs from these studies in two main aspects. One, a major part of prior work deals with scenarios in which the reward function is explicitly known by the team. Further, it is assumed that this reward function is shared by the human and the robot. In reality, however, each individual may have their own preferences about the best way of reaching the goal, which are realized as their own reward functions. This ``personal" reward function may be different from the robot's ``designer-set" reward function. 
More natural exchanges between the robot and human may not be ones in which the values are known or aligned.
In our work, we take into account this possible difference in objectives between the robot and the human and study its effect on the human's trust and team performance. We also provide results when the robot uses an adaptive interaction strategy to match its reward function to that of the human. Secondly, a majority of trust-aware decision-making research deals with robot-worker human-supervisor scenarios in which the human can choose to either observe and interrupt the robot or to focus on their own task. In our work, we are working with a robot-recommender human-decision-maker scenario in which the robot can only provide recommendations to the human on which action to select. The final say in which action to implement lies with the human. 

\subsection{Value Alignment}
\label{subsec:value-alignment}

Another area of study that has received considerable attention within the HRI community over the past few years deals with the problem of aligning the values/goals of the robot with that of its human counterpart \cite{Hadfield-Menell2016, Milli2017, Fisac2020, Yuan2022, butlin_ai_2021, moorman_impacts_2023, biyik_learning_2022, shapiro_user-agent_nodate, arnold_value_nodate}. 

Bobu et al. \cite{bobu_aligning_2024} present a brief review of literature on aligning human and robot representations and conclude by emphasizing the need for future research in this field and providing some directions for future work. Natarajan et al. \cite{natarajan_toward_2022} study the effects of adapting the driving behavior of autonomous vehicles (AVs) to the preferences of the users. Their results indicate that AVs that continuously adapt their driving style to match the dynamic preferences of users result in higher level of trust by the users. Mechergui and Sreedharan \cite{mechergui_goal_2024} propose and evaluate an interactive learning algorithm that solves the goal alignment problem when the human possesses an incorrect model of the robot's behavior.

The value alignment problem is formulated as a Cooperative Inverse Reinforcement Learning (CIRL) problem in Hadfield-Menell et al \cite{Hadfield-Menell2016}. They argue that the traditional method of training an inverse reinforcement learning agent through human demonstrations in isolation is not an optimal way to teach. They show that solutions to the CIRL problem result in behaviors for the agent that are akin to active teaching and active learning. A solution to the CIRL problem is presented in Fisac et al. \cite{Fisac2020}. This solution promotes pedagogic behavior by the human teacher, actively taking into account the learning process of the agent. The agent, on the other hand, expects this pedagogic behavior and acts pragmatically on it, resulting in efficient learning of behaviors.

Christiano et al. \cite{christiano2023deep} provide an algorithm that efficiently uses human preferences to learn robot behaviors that reach the goal states in reinforcement learning tasks. 

% Christiano et al. \cite{christiano2023deep} proposes an algorithm to learn from human preferences and shows that it can be used to ``solve'' reinforcement learning tasks in which the robot's goal is to minimize cost to reach a goal state. They show that this can be done for complex tasks within an hour of the human's time. Additionally, they show that by incorporating human preferences, the robot can learn more efficiently than using traditional deep reinforcement learning methods.

% Milli et al. \cite{Milli2017} compare a robot that completely abides by the human's literal order with a robot that instead behaves according to its estimate of the human's underlying preferences. They use simulations to compare how much more reward the human would get if the robot directly followed the human's orders vs if the robot used an estimate of the human's preferences. Their results indicate that 1) when a human is not rational, a robot should not directly obey their commands, 2) The optimal robot obeys only optimal commands from the human, and uses the estimate of the posterior mean on the reward features to drive its behavior otherwise.

There are two main differences between the work presented in this chapter and prior literature in value alignment. Firstly, as noted earlier, most prior work deals with a human-supervisor robot-worker scenario wherein the robot performs some task while the human observes it and can interrupt the operation if they want (i.e., supervisory control). We study scenarios where the robot recommends actions to the human and the final choice of action to be implemented lies with the human (i.e., decision support). Thus, in order to know when the human will accept the recommendation and when they will not, it is important for the robot to model human decision-making behavior. One way of doing this is through the level of trust of the human on the robot's recommendations. Higher the trust, higher the likelihood of the human accepting the recommendation and vice versa. Thus, we need an embedded trust dynamics and human behavior model, which is absent in most previous works in this area. Secondly, in our case, there is no \emph{true} reward function: the human and the robot have their own reward functions, and we want to see the effect of aligning/not aligning the robot's reward function with that of the human on trust and team performance. To the best of our knowledge, this is one of the first studies looking into this effect.

\section{Simulation Study}

As a first step in this research direction, we designed a study in simulation that looked at the effect of the degree of (mis)alignment of reward weights between the human and the robot on the human's trust. This section describes the details of this simulation study and presents our major results
% , and provides mathematical reasons for the observations. 

\subsection{Human-Robot Teaming Task}
\label{sec:task}
We focus on an Intelligence, Surveillance, and Reconnaissance (ISR) mission for the human-robot team. The task involves sequentially searching through $M$ search sites for potential threats. At each site, an \emph{intelligent} robot provides recommendations to the human on whether or not s/he should use an armored robot for protection. In this chapter, the term ``robot" refers to the intelligent recommender robot, unless otherwise specified. Breaching a site without the armored robot is faster as the human does not lose time in deploying the armored robot. At the same time, it is risky as the human may be harmed if a threat is encountered without protection with the armored robot. Using the armored robot is safer as it protects the human from the threat. However, it is also slower to use the armored robot as it takes time to deploy it. The objective of the team is to minimize damage to the human while also minimizing mission completion time. 

\subsection{Trust-aware Markov Decision Process}
We model the interaction between the human and the robot as they sequentially search through the town as a Trust-Aware Markov Decision Process (trust-aware MDP). It consists of states, actions, reward function, transition function, and a human trust-behavior model. 

\subsubsection{\bf States}
We define the trust of the human in the robot as the state. To quantify trust, we use the model from \cite{Bhat2022} which defines the trust $t_i$ at site $i$ to follow a Beta distribution with parameters $\alpha_i$ and $\beta_i$, i.e.,
\begin{equation}
    t_i \sim Beta(\alpha_i, \beta_i).
\end{equation}
\subsubsection{\bf Actions}
At any site $i$, the robot has two actions available: recommend to USE the armored robot or recommend to NOT USE the armored robot. So, we say that, $a^r\in\{0, 1\}$. Here, action $0$ corresponds to recommending to not use the armored robot and action $1$ to recommending to use the armored robot.
\subsubsection{\bf Rewards}
We define the rewards for each agent as a weighted sum of the health loss reward and time loss reward. The reward at site $i$ for agent $o$ is given by,
\begin{equation}
    R_i^o(a^h_i, D_i) = -w^o_hh(a_i^h,D_i)-w^o_cc(a_i^h).    
\end{equation}
Here, the weights $w_h^o$ and $w_c^o$ are the weights associated with the health loss reward and the time loss reward respectively for agent $o\in\{h, r\}$. The superscript $r$ shows that these values are for the robot and $h$ shows that the values are for the human. $a_i^h$ is the action chosen by the human at site $i$. 
% Since this is a random variable (depending on the trust and human's rewards), the reward function itself also becomes a random variable. 
$D_i$ is also a random variable which represents the presence of threat inside site $i$. $h(\cdot, \cdot)$ and $c(\cdot)$ are functions that give the health loss cost and the time loss cost respectively. In our simulation, we defined these functions as follows, $h(1, 1)=0, h(0, 1)=10, h(1, 0)=0, h(0, 0)=0$ and $c(0)=0, c(1)=10$.
\subsubsection{\bf Transition Model}
% Since the state in our MDP is the trust, our
The transition model describes the evolution of trust
% defines how trust evolves 
during the interaction. We use the reward-based performance metric from \cite{Bhat2022} to define state transitions. 
\begin{align}
    \alpha_i &= \alpha_{i-1} + P_iv^s, \\
    \beta_i &= \beta_{i-1} + (1-P_i)v^f. 
\end{align}
With the performance $P_i$ defined as, 
\begin{equation}
    P_i = \begin{cases}
        1, \text{ if } R_i^h(a^r_i)\geq R_i^h(1-a^r_i),\\
        0, \text{ otherwise.}
    \end{cases}
\end{equation}
\subsubsection{\bf Human Trust-Behavior Model}
The human behavior model encodes how a human agent responds to recommendations made by the robot. We state that the probability of the human to accept the recommendation given by the robotic agent is directly proportional to their level of trust. If the human does not accept the recommendation, s/he chooses the action that gives a higher immediate expected reward with a higher probability. More precisely, let $a^r_i$ and $a^h_i$ denote the action recommended by the robot and the action chosen by the human at site $i$ respectively. Then,
\begin{align}
\label{eq:BRD-behavior}
\begin{split}
\mathbb{P}(a^h_i=a|a^r_i = a) &= t_i + (1-t_i)p^a_i,\\
\mathbb{P}(a^h_i=1-a|a^r_i = a) &= (1-t_i)(1 - p^a_i).
\end{split}
\end{align}
Here $t_i$ is the trust level at the $i^{th}$ site and $p^a_i$ is given by, 
\begin{equation}
    \label{eq:p-a}
    p^a_i = \frac{\exp(\kappa E[R^h_i(a)])}{\exp(\kappa E[R^h_i(a)]) + \exp(\kappa E[R^h_i(1-a)])} .   
\end{equation}
Here, the superscript $a$ can be either $0$ or $1$ corresponding to the two actions. $\kappa$ controls the effect of the rewards on the probability $p^a_i$. In human behavior modeling literature, this is often referred to as the rationality coefficient \cite{BAKER2014, Fisac2020, Yuan2022}. The higher the value of $\kappa$, the more likely it is that the human will choose the action with the higher reward. On the other hand, $\kappa=0$ represents a uniformly random human, one who chooses each action with equal probability. 
We call the resulting model the \emph{bounded rationality disuse} model since it combines concepts from both models.

\subsubsection{\bf Value Iteration}
\label{sec:ValueIteration}
The robot solves the trust-aware MDP via value iteration. It selects the action by maximizing the expected reward at the current step summed together with a discounted value of the next state at the next stage.

\begin{equation}
\label{eq:q-value-function}
Q_i(s_i, a) = E\left[R_i^r(a)\right] + \sum_{s_{i+1}\in S}\gamma\mathbb{P}(s_{i+1}|s_i,a)V_{i+1}(s_{i+1}).
\end{equation}
\begin{equation}
\label{eq:value-function-max}
V_i(s_i) = \max_a{Q_i(s_i, a)}.
\end{equation}

At the final stage, the action that gives the maximum immediate expected reward is chosen.

\begin{equation}
\label{eq:value-function-last}
V_N(s_N) = \max_a{E\left[R^r_N(a)\right]}.
\end{equation}

\subsubsection{\bf Bayesian Inverse Reinforcement Learning}
\label{subsec:BayesianIRL}
The robot uses a Bayesian Inverse Reinforcement Learning \cite{Ramachandran2007} framework with the assumed human trust-behavior model to learn the human's preferences between saving health and saving time. The robot maintains a belief $b_i(w_h^h)$ over the possible health reward weights of the human. Here, the subscript $i$ denotes that this is the robot's belief just before the $i^{th}$ site is searched. After observing the action selected by the human, the robot updates this belief using Bayes rule. 

If at site $i$, we observe that the human chooses the recommended action, we update the belief according to, 
\begin{align}
    \label{eq:IRL-follow}
    b_{i+1}(w) &\propto \mathbb{P}(a^h_i=a|w^h_h=w,a^r_i=a,t_i)b_i(w)\nonumber\\
               &\propto (t_i + (1-t_i)p^a_i)b_i(w).
\end{align}
If, at site $i$, we observe that the human chooses the action opposite to the one recommended, we update the belief according to,
\begin{align}
    \label{eq:IRL-not-follow}
    b_{i+1}(w) &\propto \mathbb{P}(a^h_i=1-a|w^h_h=w,a^r_i=a,t_i)b_i(w)\nonumber\\
               &\propto (1 - t_i)(1 - p^a_i)b_i(w).
\end{align}
Here, $p^a_i$ is computed according to Eq. \ref{eq:p-a}, substituting weight $w$ in the expected reward computation. We use the mean of the belief distribution as a proxy for the human's health weight. We always set $w_c^h=1-w_h^h$.

\subsubsection{Threats and Threat Levels}
We follow the following strategy to set the threat levels at the search sites. First, a prior probability of threat presence $d$ is fixed across all sites. Then, at each site independently, threat presence is determined by a Bernoulli distribution with $d$ as the parameter. The drone gets an updated threat level after scanning a site. This updated threat level is generated by a beta distribution with a peak at $0.1$ for when a threat is not present and a peak at $0.9$ for when a threat is present inside the site. 
\subsubsection{Simulating Human Decision and Feedback}
We sample from the set of trust parameters $\Theta$ that was generated from a previous study \cite{Bhat2022} to get $\theta = (\alpha_0, \beta_0, v^s,v^f)$ for the simulated human. These parameters are unknown to the robot and it estimates them via maximum likelihood estimation (for details, refer to \cite{Bhat2022}). The human also has reward weights $w^h_h, w^h_c(:=1-w^h_h)$ which are unknown to the robot. The human gives trust feedback by sampling from the beta distribution trust update model, using the observed rewards associated with the recommendation to judge the performance of the robot. The human chooses his/her action using the bounded rationality disuse model with $\kappa = 1$.

\subsection{Results - Simulation Study}
\label{sec:results-simulation}
This section provides details about the major results from this simulation study.
\subsubsection{Regions in End-of-mission Trust}
\label{sec:regions}
\begin{figure}[ht]
\centering
\subcaptionbox{$d=0.3$ \label{fig:Region-0.3}}[0.45\textwidth]{\includegraphics[width=\linewidth]{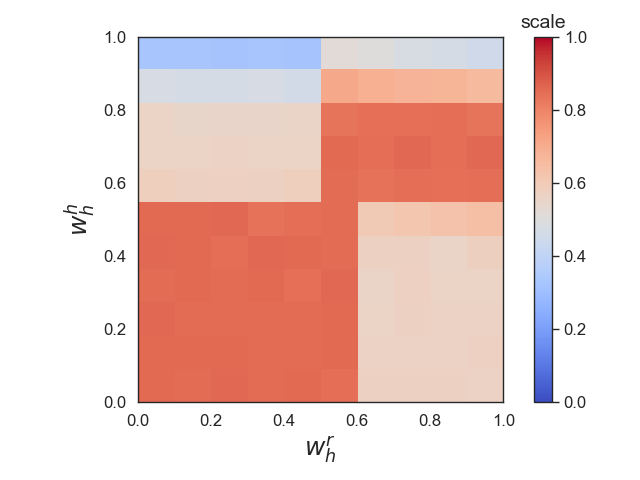}}
\hfil
\subcaptionbox{$d=0.7$ \label{fig:Region-0.7}}[0.45\textwidth]{\includegraphics[width=\linewidth]{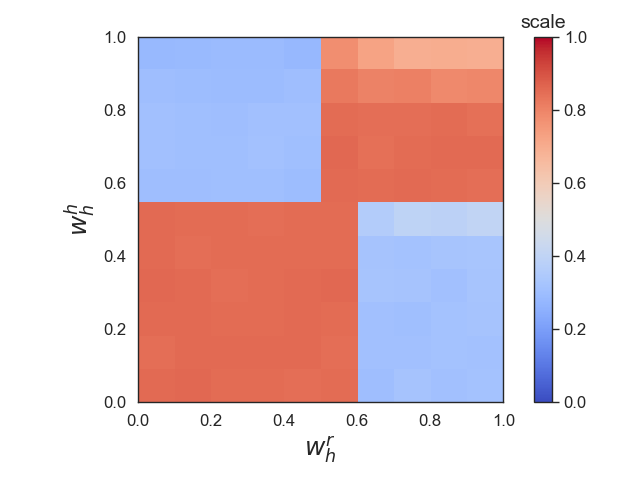}}
% \subfigure[$d=0.7$]{
% \includegraphics[width=0.35\linewidth]{images/Simulation/Regions/BoundedRational/regions-2d-0.7.png}
% \label{fig::Region-0.7}}

\caption{The observed regions in end-of-mission trust as a function of the health weights of the human $w^h_h$ and the robot $w^r_h$. The figure on the left is the simulation result when there is a relatively low chance of threat presence at any search site $(d=0.3)$. The figure on the right is when there is a higher chance of threat presence at any search site $(d=0.7)$. 
}
\label{fig:regions-in-trust}
\end{figure}

We observed four distinct regions in the end-of-mission trust as a function of the reward weights of the human and the robot (Fig. \ref{fig:regions-in-trust}). If a risk-averse human is paired with a risk-taking robot, the end-of-mission trust can get very low (the region in the top-left of the plots in Fig. \ref{fig:regions-in-trust}) and vice versa. This is especially true when the risk level is higher, as can be seen in the higher difference in trust levels when the values align and when they don't in Fig. \ref{fig:Region-0.7} compared to the same in Fig. \ref{fig:Region-0.3}. In the figure on the left, the probability of threat being present in any site was $0.3$ while the same was $0.7$ in the figure on the right. 

% \begin{figure}[ht]
% \centering
% \subfigure[$d=0.3$]{
% \includegraphics[width=0.35\linewidth]{images/Simulation/Regions/BoundedRational/regions-2d-0.3.png}
% \label{fig::Region-0.3}}
% \hfil
% \subfigure[$d=0.7$]{
% \includegraphics[width=0.35\linewidth]{images/Simulation/Regions/BoundedRational/regions-2d-0.7.png}
% \label{fig::Region-0.7}}

% \caption{The observed regions in end-of-mission trust as a function of the health weights of the human $w^h_h$ and the robot $w^r_h$. The figure on the left is the simulation result when there is a relatively low chance of threat presence at any search site $(d=0.3)$. The figure on the right is when there is a higher chance of threat presence at any search site $(d=0.7)$. 
% }
% \label{fig:regions-in-trust}
% \end{figure}

\subsubsection{Effect of Threat Level}
\begin{figure}[ht]
\centering
\subcaptionbox{Region 1 \label{fig:region1-threat}}[0.4\linewidth]{\includegraphics[width=\linewidth]{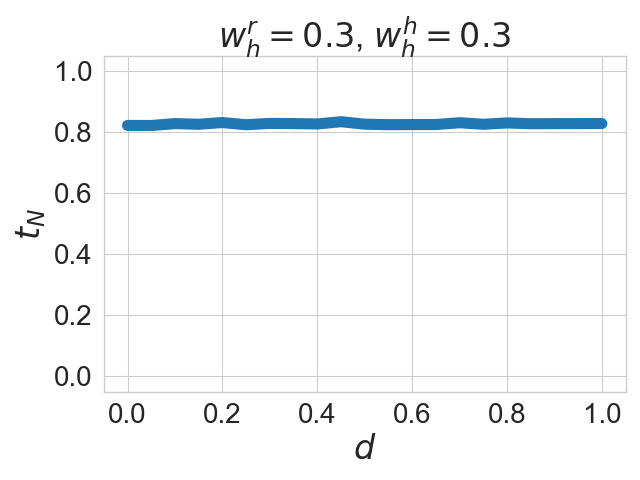}}
\hfil
\subcaptionbox{Region 2 \label{fig:region2-threat}}[0.4\linewidth]{\includegraphics[width=\linewidth]{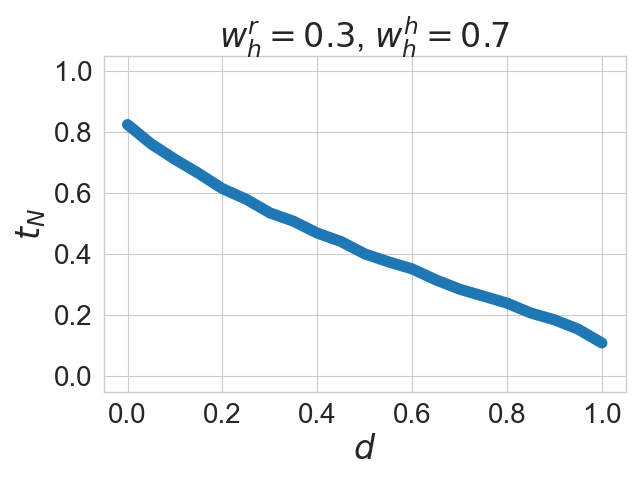}}
\hfil
\subcaptionbox{Region 3 \label{fig:region3-threat}}[0.4\linewidth]{\includegraphics[width=\linewidth]{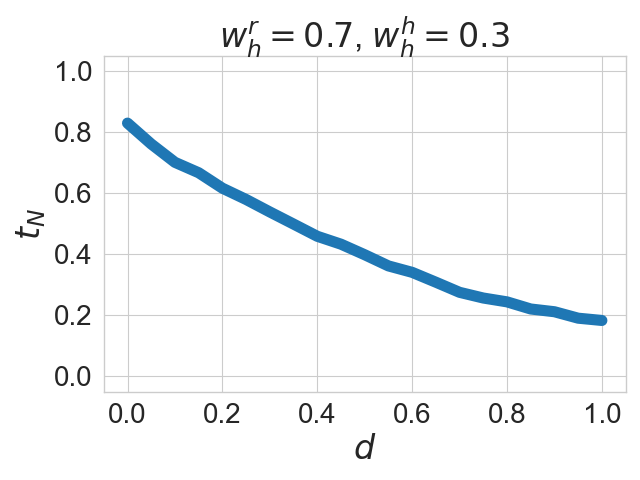}}
\hfil
\subcaptionbox{Region 4 \label{fig:region4-threat}}[0.4\linewidth]{\includegraphics[width=\linewidth]{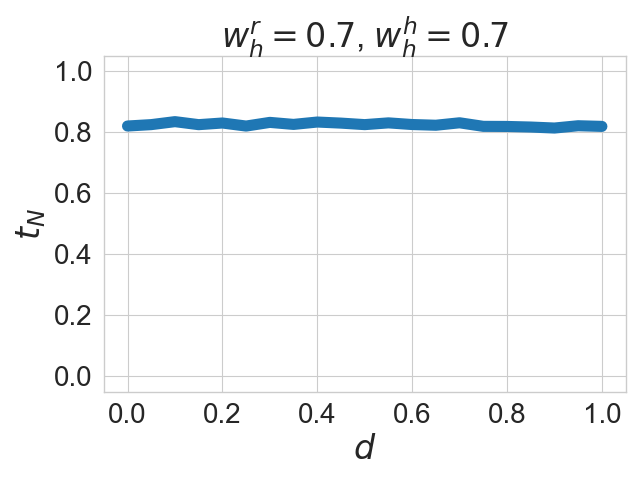}}
\caption{The effect of the prior probability of threat presence in any house $d$ on the end-of-mission trust, after fixing a trust region.}
\label{fig:effect-of-threat-BRD}
\end{figure}

Since the results from the last section show some potential effects of the level of threat associated with the mission on the end-of-mission trust, we ran another set of simulations, this time, fixing the reward weights of the human and the robot (i.e. the trust region) and varying the level of threat. Fig. \ref{fig:effect-of-threat-BRD} shows plots of the end-of-mission trust as a function of the prior threat level $d$ for the 4 regions of trust. Further, it is seen that if the values are roughly aligned, trust increases. If the values are not aligned, trust is still high when the overall risk is low (at low threat levels $d$). However, the human starts to lose trust in the robot if the values are misaligned and the risk level is high.

\begin{figure}
    \centering
    \includegraphics[width=0.4\linewidth]{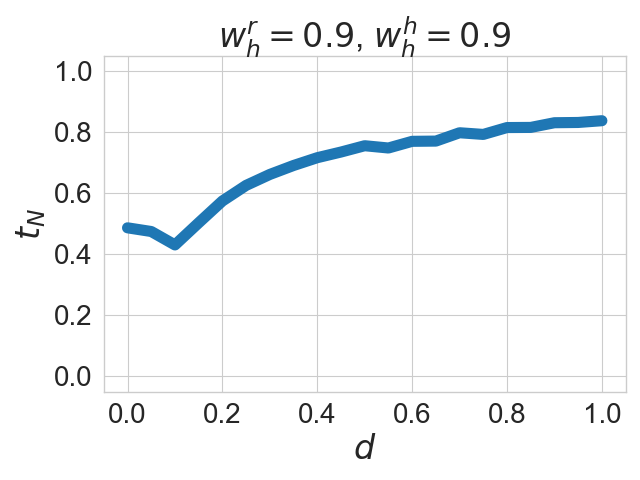}
    \caption{The effect of prior probability of threat presence in any house $d$ on the end-of-mission trust when the human and the robot are both extremely risk-averse}
    \label{fig:effect-of-threat-BRD-extreme}
\end{figure}

Fig. \ref{fig:effect-of-threat-BRD-extreme} shows an example when both the human and the robot are extremely risk-averse. In this case, although there is alignment in the reward weights of the human and the robot, the level of trust is generally very low.
% It however does show that trust tends to decrease even when there is value alignment in a broader sense in that both the human and the robot care about protection from loss of health, but the robot is overly cautious. 
This is because the robot tries to maximize expected reward, but the human is judging performance through observed rewards. At lower risk levels, it would be better to be cautious in the long term, but it is more likely to get better rewards in the short term by being risky. Thus, it seems to the human that the robot is overly cautious, resulting in a decrease in trust. 

\subsubsection{Effect of Adapting to the Human's Rewards}
We can update the reward weights of the robot at each time step by using the distribution on the reward weights of the human maintained by the robot (see Sec. \ref{subsec:BayesianIRL}). This results in an ``adaptive" interaction strategy in which the robot adapts to human values throughout their interaction. We ran simulations when the robot followed such a strategy, starting with a uniform distribution on the reward weights. We can then compare this adaptive strategy with a non-adaptive strategy of the robot with $w^r_h=0.5$ (the mean of the initial uniform distribution) to see the effect of adapting to human preferences. 

\begin{figure}[ht]
    \centering
    \subcaptionbox{$d=0.3$ \label{fig:adaptive-threat-0.3}}[0.45\textwidth]{\includegraphics[width=\linewidth]{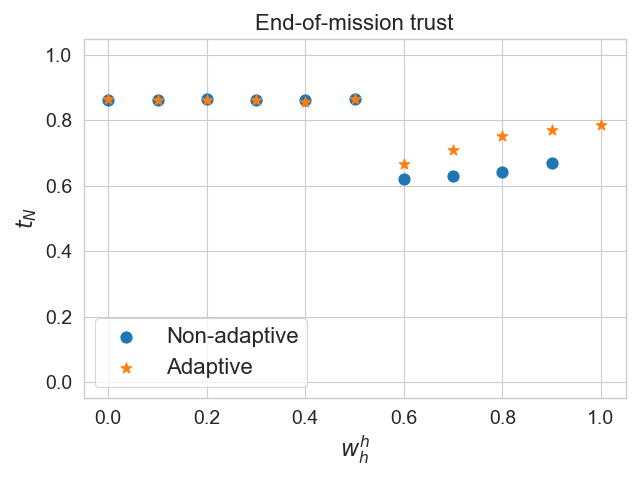}}
    \hfil
    \subcaptionbox{$d=0.7$ \label{fig:adaptive-threat-0.7}}[0.45\textwidth]{\includegraphics[width=\linewidth]{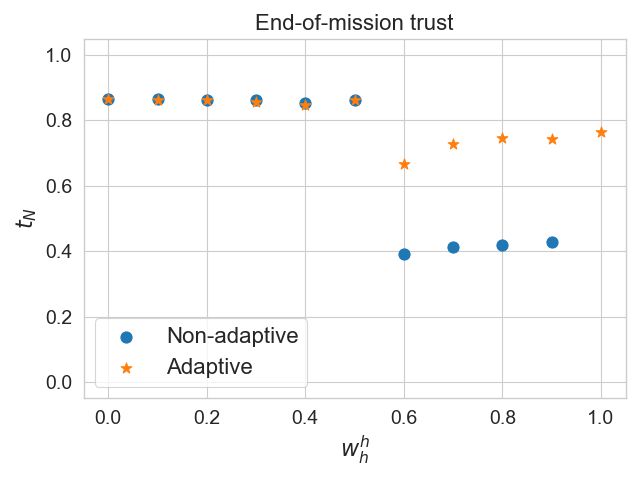}}
    \caption{Comparing the adaptive strategy with the non-adaptive strategy in the end-of-mission trust feedback given by the simulated human for two different levels of threat. The non-adaptive strategy sets the robot's health reward weight to $0.5$}
    \label{fig:adaptive-effect-of-weights}
\end{figure}

Fig. \ref{fig:adaptive-effect-of-weights} shows the effect of the human's health reward weight on the end-of-mission trust reported by the human at two threat levels: low $(0.3)$ and high $(0.7)$. As can be seen, below $w_h^h<0.5$, there is no clear difference between trust on the non-adaptive strategy and that on the adaptive strategy. This is because at $w^r_h=0.5$, the robot has no preference between losing health and losing time. However, in expectation, not using the RARV is always better than using the RARV. Thus, the recommendation matches the human's preferred action (since the human also does not care about losing health when $w^h_h<0.5$). As a result, trust is high for both non-adaptive and adaptive strategies. On the other hand, however, when $w_h^h>0.5$, the human prefers to save health over saving time. Thus, the risky recommendations of the non-adaptive strategy reduce trust while the adaptive strategy is able to learn this preference and is able to maintain the human's trust.

\begin{figure}[ht]
    \centering
    \subcaptionbox{$w_h^h=0.3$ \label{fig:adaptive-weight-0.3}}[0.45\linewidth]{\includegraphics[width=\linewidth]{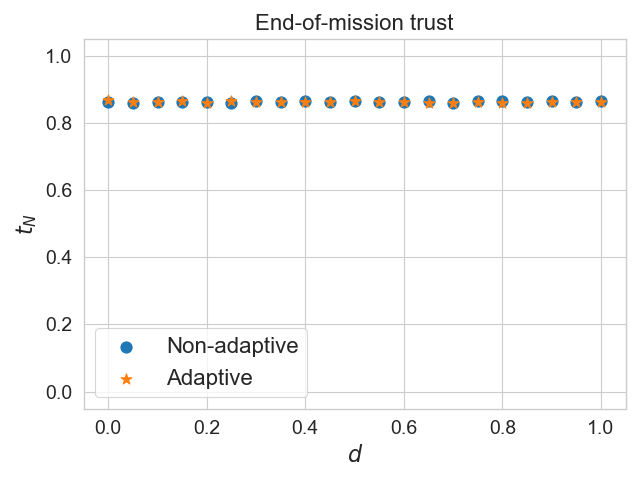}}
    \hfil
    \subcaptionbox{$w_h^h=0.7$ \label{fig:adaptive-weight-0.7}}[0.45\linewidth]{\includegraphics[width=\linewidth]{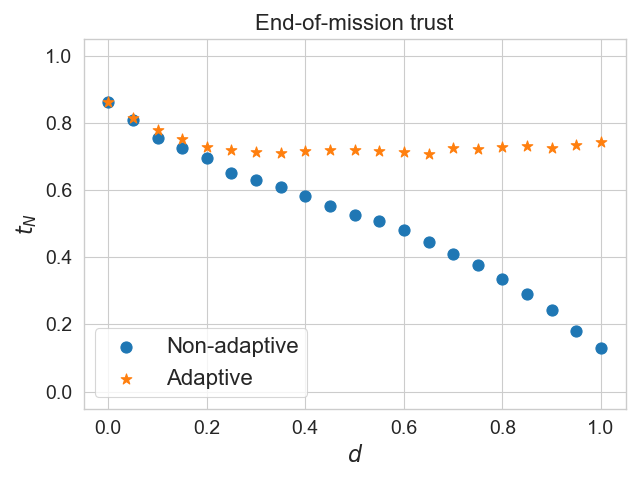}}
    \caption{Comparing the adaptive strategy with the non-adaptive strategy in the end-of-mission trust feedback given by the simulated human for two different levels of threat. The non-adaptive strategy sets the robot's health reward weight to $0.5$}
    \label{fig:adaptive-effect-of-threat}
\end{figure}

Fig. \ref{fig:adaptive-effect-of-threat} compares the two strategies for two different human preferences (preferring to lose health $w^h_h=0.3$ and preferring to lose time $w^h_h=0.7$) as the level of threat is changed. Again, we see similar behavior that when the human prefers to lose health, both strategies result in the same level of trust of the human on the robot. On the other hand, when the human prefers to lose time, the non-adaptive strategy's risky recommendations lead to a loss in trust as the level of threat increases. Again, the adaptive strategy is able to learn this preferences resulting in a much higher level of the human's trust, irrespective of the level of threat. 

\subsection{Discussion - Simulation Study}
A key takeaway from our results is that, under the bounded rationality disuse model value alignment is, in general, good for trust. However, at low risk settings, value alignment may not be as important for gaining trust. This conclusion is somewhat in line with the finding of \cite{Lyons2012}, wherein the authors found increased reliance on automation in high risk situations compared to low risk situations. Parkhe and Stewart \cite{Parkhe2000} claim that trust is only necessary when the losses can exceed potential gains (i.e. when the risk is high). We take it a step further by saying that value alignment is only necessary for trust when the risks are high. Further, since humans are generally more concerned with observed outcomes rather than expected rewards, the boundaries of the regions can result in lower end-of-mission trust despite having values aligned between the human and the robot. 

% This study has several limitations. Firstly, we believe that the binary trade-off between losing health and losing time (in other words having two features in the reward function) is a major reason why we see such clear regions in the end-of-mission trust. If the reward function has more components in it, we may not see similar behavior of trust. Secondly, we demonstrate our framework in a binary action scenario: using or not using an armored robot. We do note, however, that our bounded rationality disuse trust-behavior model, the value iteration formulation, and the inverse reinforcement learning framework can easily be extended to the case where there are multiple actions to choose from. However, changing the performance measure for such a multiple-action scenario could prove challenging, especially due to the existence of satisficing \cite{Simon1956}. This will be an interesting direction for future study.
% % In a `robot recommender, human follower' scenario, we do expect that the action sets would be the same for the robot and the human. 
% Lastly, we have assumed that the robot's scanner is perfectly reliable, giving a high chance of threat presence when a threat is actually present in the search site and vice versa. 

In light of our results, we claim that completely aligning the robot's values to the human's may not always be the best, especially when the human is extremely risk-averse. This may result in loss of performance \emph{and} loss of trust. On the other hand, not aligning the values at all will certainly lead to a loss in trust, especially in high risk situations. Thus, we postulate that there may be merit in looking for a middle ground that trades off between fully aligning the values and not aligning them at all. This could be a very interesting direction for future research. 

\section{Empirical Study}
This section provides details on the human-subjects experiment carried out to empirically validate the results of the simulation study. An extension of this study is presented in \cite{bhat_effect_2024, bhat_evaluating_2024}. Interested readers are encouraged to refer to these studies for further details.
% We conducted two studies, with the difference being the information available to the robot about the general population's preferences between saving the soldier's health and saving mission time. In study 1, the robot has a data-driven prior on these preferences while in study 2, the robot starts its learning with an uninformed prior. 

\subsection{Testbed}
\begin{figure}[ht]
\centering
\subcaptionbox{The recommendation interface \label{fig:rec-interface}}[0.45\linewidth]{\includegraphics[width=\linewidth]{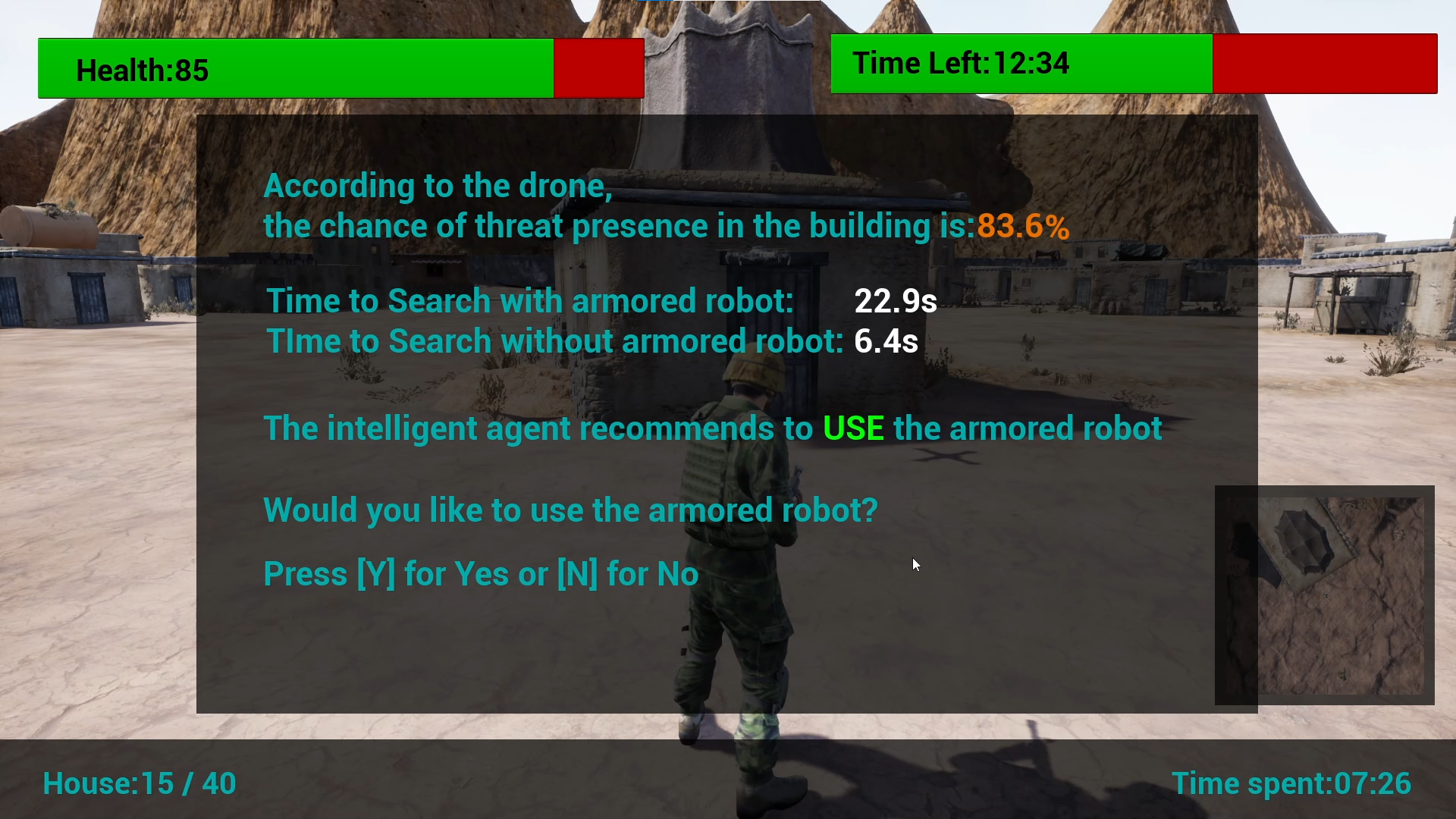}}
\hfil
\subcaptionbox{Trust feedback slider \label{fig:feedback-slider}}[0.45\linewidth]{\includegraphics[width=\linewidth]{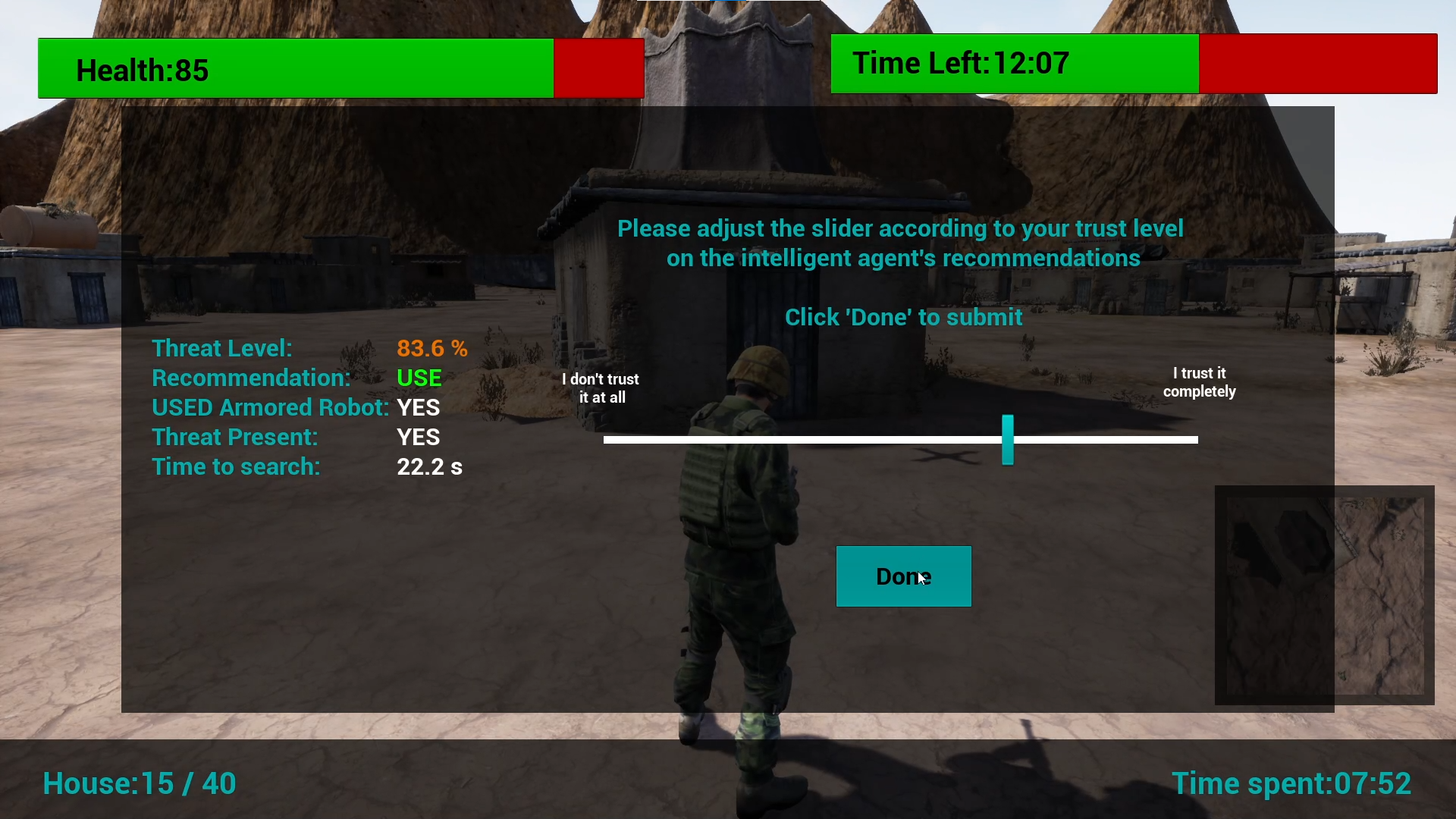}}
\caption{The recommendation interface showing the information about the threat level and the recommendation by the intelligent agent. Also shown, the trust feedback slider used to get feedback from the participants after every search site. The mission timer is paused when the slider is shown to let the participants take their time in adjusting their trust.}
\label{fig:testbed}
\end{figure}

We developed a 3D testbed using the Unreal Engine game development platform. Within the testbed, a human is paired with an autonomous drone and an armored robot to complete the reconnaissance mission described in Sec. \ref{sec:task}. Fig. \ref{fig:rec-interface} shows a screenshot of the recommendation interface used by the intelligent agent. It displays the threat level indicated by the drone, the average time to search a site with and without the Robotic Armored Rescue Vehicle (RARV) and the system's recommendation. The participant is then free to select an action by pressing the associated key on the keyboard. The participants are, at all times, displayed the time left to complete the mission and the current health level of the human. They are also displayed the index of the current search site. They are given a total of $25$ minutes to complete searching a total of $40$ sites. Each time the human encounters a threat without protection from the RARV, they lose $5$ points of health. Deploying the RARV takes approximately $15$ seconds. After exiting a search site, the participants are asked to report their trust on the recommendation system on a slider (shown in Fig. \ref{fig:feedback-slider}). They are shown the drone's assessed level of threat, the recommendation given by the system, the action selected by the participant, the ground truth of threat presence, and the time it took them to search the site on the feedback dialog to aid with recall. The mission timer is stopped when the participants are shown the feedback slider. This is to let the participants take their time to assess and report their trust accurately. The threats and threat levels the drone gets after scanning a site are randomly generated while ensuring that there are threats in $23$ out of the $40$ sites, corresponding to a prior threat level $d=0.575$.

\subsection{Conditions}
We designed three interaction strategies for the intelligent agent:
\begin{itemize}
    \item \textbf{Non-learner:} The intelligent agent does not learn the reward weights of the human. It assumes that the human shares the same reward weights as itself and uses these for recommendation success assessment, trust updating, human behavior modeling, and MDP optimization. 
    \item \textbf{Non-adaptive learner:} The intelligent agent learns personalized reward weights during interaction with each human. It only uses these learned weights for recommendation success assessment, trust updating, and human behavior modeling. It still optimizes the MDP with its own fixed reward weights. 
    \item \textbf{Adaptive learner:} The intelligent agent learns personalized reward weights for each human. It then adopts this learnt reward weights as its own.
\end{itemize}

We employed a within-subjects design. Each participant completed three missions. In each mission, they interacted with an intelligent agent using one of the interaction strategies. They were not informed which interaction strategy the robot was using. They were only informed of their dual goal of minimizing damage to the soldier while also completing the mission as quickly as possible. A $3 \times 3$ Latin Square design was used to minimize any learning effects.

\subsection{Participants}
We collected data from a total of $24$ participants (Age: Mean $21.4$ years, \textit{SD} $2.3$, 12 Female). All participants were students from the University of Michigan.

% , $30$ of which participated in experiment $1$ (Age: Mean $22.6$ years, \textit{SD} $3.6$, 14 Female) and $24$ participated in experiment $2$ (Age: Mean $21.4$ years, \textit{SD} $2.3$, 12 Female). All participants were students from the University of Michigan.

\subsection{Measures}
\subsubsection{Pre-experiment Measures}
Prior to the experiment, participants completed a demographic survey indicating their age, gender, academic department, nationality, frequency and skill of playing video games, and familiarity with AI/ML algorithms. Participants also filled in questionnaires about their personality, propensity to trust autonomy, and type of decision-making style. 

\subsubsection{Pre-mission Measures}Before each of the three missions, participants rated their preferences. 

\begin{itemize}
    \item \textbf{Task Preference:} Before the beginning of each mission, we ask the participants to rate their preference between saving health and saving time by moving a slider between these two objectives, showing their relative importance.
\end{itemize}

\subsubsection{In Experiment Measures}
After each site's search was completed (i.e., every trial), the participants were asked to report their level of trust in the intelligent agent, $t_{i}$ (see fig. \ref{fig:feedback-slider} for the exact question asked). The slider values were between 0 and 100 with a stepsize of 2 points. Additionally, for every trial, we measured whether the participants agreed with the intelligent agent's recommendation. With the trial-based data, we measured the following:

\begin{itemize}
    \item \textbf{Average Trust:} This was calculated as the empirical mean trust $\frac{1}{M}\sum_{i=1}^M t_i$.
    % \item \textbf{End-of-mission Trust:} This was the participant's self-reported trust after the last trial, $t_M$.
   \item \textbf{Number of Agreements:} This was computed as the number of times the participant chose the recommended action.
    \end{itemize}
  Note that $M=40$ is the number of sites in a mission.

\subsubsection{Post-mission Measures} After every mission, participants completed survey assessing the following items. 
\begin{itemize}
    \item \textbf{Post-mission trust questionnaire:} Measured using Muir's trust questionnaire \cite{Muir1996}. It has 9 questions, each with a slider with a range between 0 and 100. 
    \item \textbf{Post-mission Reliance Intentions:} Measured using the scale developed by Lyons and Guznov \cite{Lyons2019}. We used 6 of the 10 items that were relevant for this task. Each item was rated on a 7-point Likert scale.
    \item \textbf{Workload:} Workload was measured using the NASA Task Load Index (NASA TLX) scale \cite{Hart1988, Lu2019, luo_workload_2021}. We used 5 of the 6 dimensions as our experiment involved minimal physical demand. Each item was measured using a slider ranging from very low to very high. 
    \item \textbf{Performance:} We computed the team performance by a weighted sum of the percentage health remaining of the soldier and the percentage time remaining in the mission. The weights used were the participants rated preferences at the beginning of the mission.

   \begin{equation}
       \text{Performance} = \hat{w}^h_h\cdot (\% h_M) + \hat{w}^h_c \cdot (100 - \%c_M).
   \end{equation}

    where $\hat{w}^h_h$ and $\hat{w}^h_c:=1-\hat{w}^h_h$ are the reported preferences by the participant before beginning the mission, $\%h_M$ is the percent health remaining and $\%c_M$ is the percent time spent at the end of the mission. This metric allows us to convert the two conflicting objectives with different units of measurement into one unified scale. The higher the value, the better the team performance. 

\end{itemize}

\subsection{Results - Empirical Study}
This section summarizes our results and discusses the implications (see Table \ref{tab:results} for an overview). Repeated measures analyses of variance (ANOVAs) were conducted to compare the three interaction strategies. Greenhouse-Geisser corrections to the degrees of freedom were made for the measures that failed Mauchly's test of sphericity. 

% In Experiment 1, we initiated the IRL learning algorithm with an informed prior. In Experiment 2, the learning algorithm started with an uninformed prior.

\begin{table}[ht]
    \centering
    \caption{Mean $\pm$ standard deviation (SD) of measures for the interaction strategies}
    \vspace{3mm}
    \label{tab:results}
    \begin{tabular}{|lccc|}
\hline
\multicolumn{1}{|l|}{}                                           & \multicolumn{1}{c|}{\textbf{Non-learner}}       & \multicolumn{1}{c|}{\textbf{Non-adaptive learner}} & \textbf{Adaptive-learner}  \\ \hline
\multicolumn{1}{|l|}{Average trust $\frac{1}{M}\sum_{i=1}^Mt_i$ $^\ast$} & \multicolumn{1}{c|}{$0.42 \pm 0.20$}   & \multicolumn{1}{c|}{$0.45 \pm 0.22$}      & $0.65 \pm 0.20$   \\
\multicolumn{1}{|l|}{End-of-mission trust $t_M$ $^\ast$}                   & \multicolumn{1}{c|}{$0.33 \pm 0.30$}   & \multicolumn{1}{c|}{$0.35 \pm 0.29$}      & $0.64 \pm 0.30$   \\
\multicolumn{1}{|l|}{Muir's trust questionnaire $^\ast$}                 & \multicolumn{1}{c|}{$34.51\pm 19.79$}  & \multicolumn{1}{c|}{$34.97 \pm 17.41$}    & $60.57\pm 23.71$  \\
\multicolumn{1}{|l|}{Number of agreements $^\ast$}                       & \multicolumn{1}{c|}{$28.17 \pm 5.49$}  & \multicolumn{1}{c|}{$28.67 \pm 4.65$}     & $35.62 \pm 2.93$  \\

\multicolumn{1}{|l|}{Reliance intentions scale $^\ast$}                  & \multicolumn{1}{c|}{$2.16 \pm 1.22$}   & \multicolumn{1}{c|}{$2.38 \pm 1.16$}      & $3.58 \pm 1.32$   \\
\multicolumn{1}{|l|}{Workload $^\ast$}                                   & \multicolumn{1}{c|}{$38.18 \pm 13.79$} & \multicolumn{1}{c|}{$39.82\pm 14.94$}     & $31.13 \pm 10.64$ \\
\multicolumn{1}{|l|}{Performance}                                & \multicolumn{1}{c|}{$45.86 \pm 16.20$} & \multicolumn{1}{c|}{$49.11 \pm 14.25$}    & $51.60 \pm 18.68$ \\ \hline
\multicolumn{4}{l}{\footnotesize $\ast - p<0.05$}
\end{tabular}
\end{table}

\subsection{Trust, Agreement, and Reliance Intention }

% \subsubsection{Experiment 1: with informed prior}
% We observed no significant difference between the three strategies in average trust $(F(2, 58) = 0.308,~ p = 0.736)$, end-of-mission trust $(F(2, 58) = 1.192,~ p = 0.311)$, and the  Muir's trust scale ($F(2, 58) = 1.550, p = 0.221$). 

% Additionally, there was no significant difference in the number of agreements $(F(2, 58) = 0.755,~ p = 0.475)$ across the three strategies. However, there was a significant difference in reliance intentions $(F(1.543, 44.737), ~p=0.031)$ (Fig. \ref{fig:reliance-intentions-2a}). Pairwise comparisons with Bonferroni adjustments revealed a lower intent to rely on the adaptive learner strategy than the non-learner strategy $(p=0.012)$. 

% \begin{figure}[h]
%     \centering
%     \includegraphics[width=0.45\columnwidth]{images/HumanSubject/barcharts/phase2a/trust/reliance-intentions.png}
%     \caption{Exp 1 -- Post-mission reliance intentions}
%     \label{fig:reliance-intentions-2a}
% \end{figure}

% \subsubsection{Experiment 2: with uninformed prior}

\begin{figure}[ht]
\centering
\subcaptionbox{Average Trust $\frac{1}{M}\sum_{i=1}^Mt_i$ \label{fig:avg-trust-2b}}[0.45\linewidth]{\includegraphics[width=\linewidth]{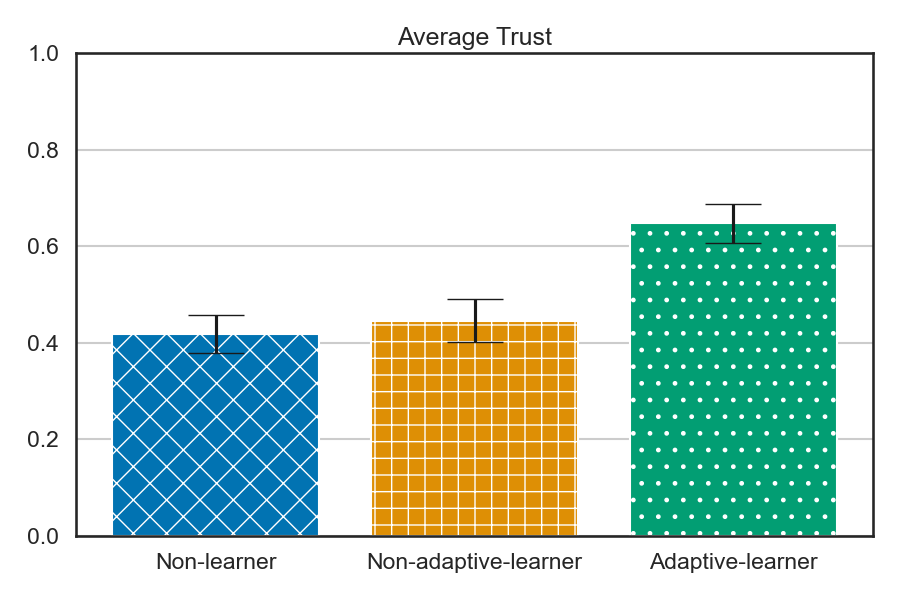}}
\hfil
\subcaptionbox{Subjective Trust - Muir's scale \label{fig:trust-muir-2b}}[0.45\linewidth]{\includegraphics[width=\linewidth]{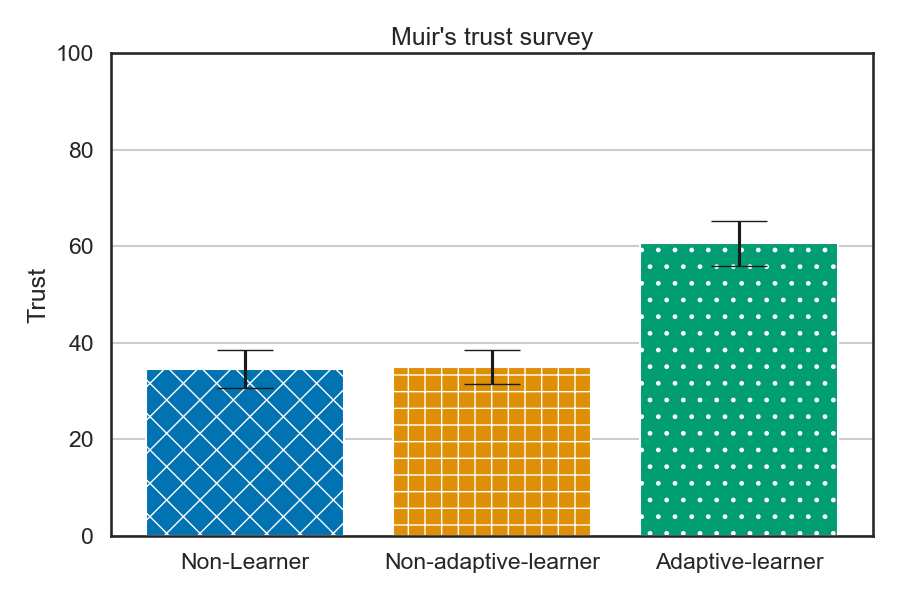}}
\subcaptionbox{Number of Agreements \label{fig:agreements-2b}}[0.45\linewidth]{\includegraphics[width=\linewidth]{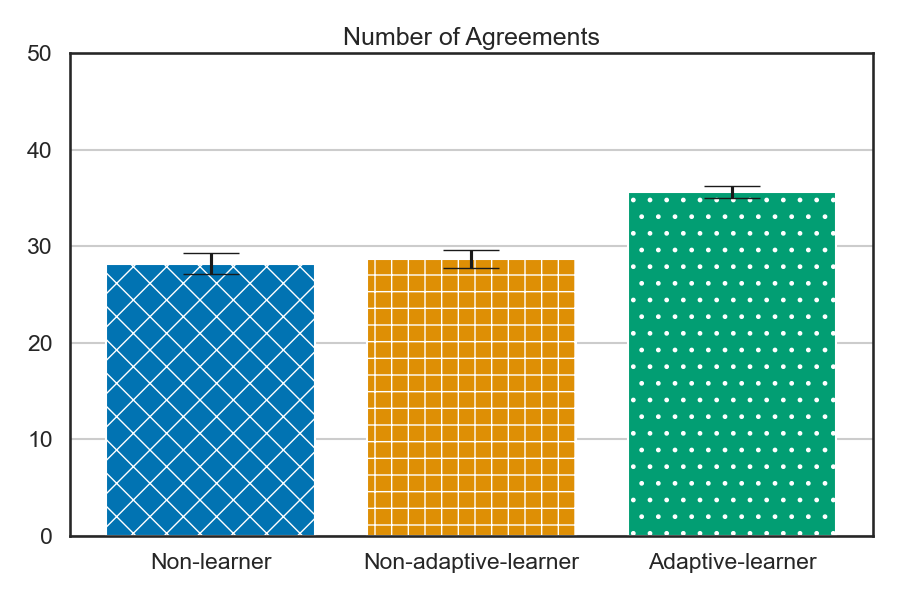}}
\hfil
\subcaptionbox{Reliance Intentions \label{fig:reliance-intentions-2b}}[0.45\linewidth]{\includegraphics[width=\linewidth]{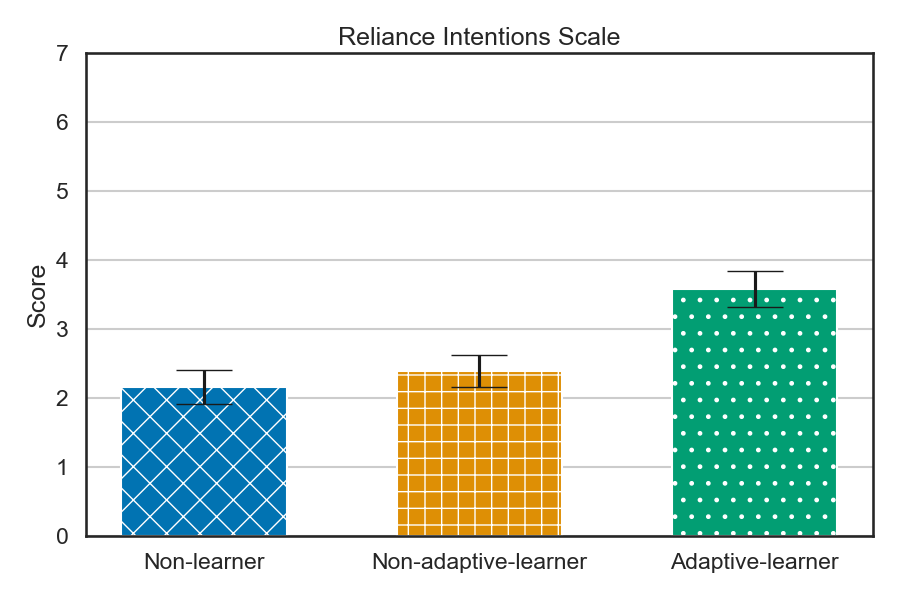}}
\caption{Comparing the human's trust and reliance intentions on the robot when the robot follows different interaction strategies}
% for the second study (when the robot starts learning from scratch)}
\label{fig:phase2b-trust}
\end{figure}

Figs. \ref{fig:avg-trust-2b} and \ref{fig:trust-muir-2b} show the comparisons of the three strategies in trust. Repeated measures ANOVA revealed significant differences between the three strategies in average trust $(F(2, 46)=14.161, ~p < 0.001)$ and Muir's trust scale $(F(1.586,\\ 36.473) = 16.3, ~p < 0.001)$.

Pairwise comparisons with Bonferroni adjustments revealed that the adaptive-learner strategy led to higher average trust and post-mission trust compared to the non-learner strategy ($p<0.001$ and $p<0.001$ respectively) and compared to the non-adaptive learner strategy ($p=0.003$ and $p<0.001$ respectively). 

Regarding the number of agreements (Fig. \ref{fig:agreements-2b}), there was a significant difference among the three strategies $(F(1.584, 36.435) = 25.829, ~p < 0.001)$. Post-hoc analysis showed that there was a significant difference between the non-learner and the adaptive-learner strategies $(p<0.001)$ and between the non-adaptive-learner and adaptive-learner strategies $(p<0.001)$. 

Comparing reliance intentions (Fig. \ref{fig:reliance-intentions-2b}), there was a significant difference between the three strategies ($F(2, 46) = 13.691, p < 0.001$), with the adaptive-learner strategy rated higher than the non-learner strategy ($p<0.001$) and the non-adaptive-learner strategy ($p=0.004$).

\subsection{Performance}

% \subsubsection{Experiment 1: with informed prior}

% Even though there seemed to be a decreasing performance trend from non-learner to non-adaptive learner and to adaptive learner (i.e., $61.47 \pm 18.12$, $60.25\pm 17.03$, and $55.83\pm 20.39$) the trend did not reach statistical significance $(F(2,58)=2.067,~p=0.136)$. 

% \subsubsection{Experiment 2: with uninformed prior}

There seemed to be an upward trend from non-learner to non-adaptive learner and to adaptive-learner. Unfortunately, it did not reach significance. 

\subsection{Workload}

% \subsubsection{Experiment 1: with informed prior}
% Comparing the average workload across the three interaction strategies showed no significant difference $(F(2, 58) = 2.634, ~p = 0.089)$ (Table \ref{tab:results}). Additionally, repeated measures ANOVAs did not reveal any significant differences between the three strategies in any of the dimensions.

\begin{figure}[h]
    \centering
    \includegraphics[width=0.45\columnwidth]{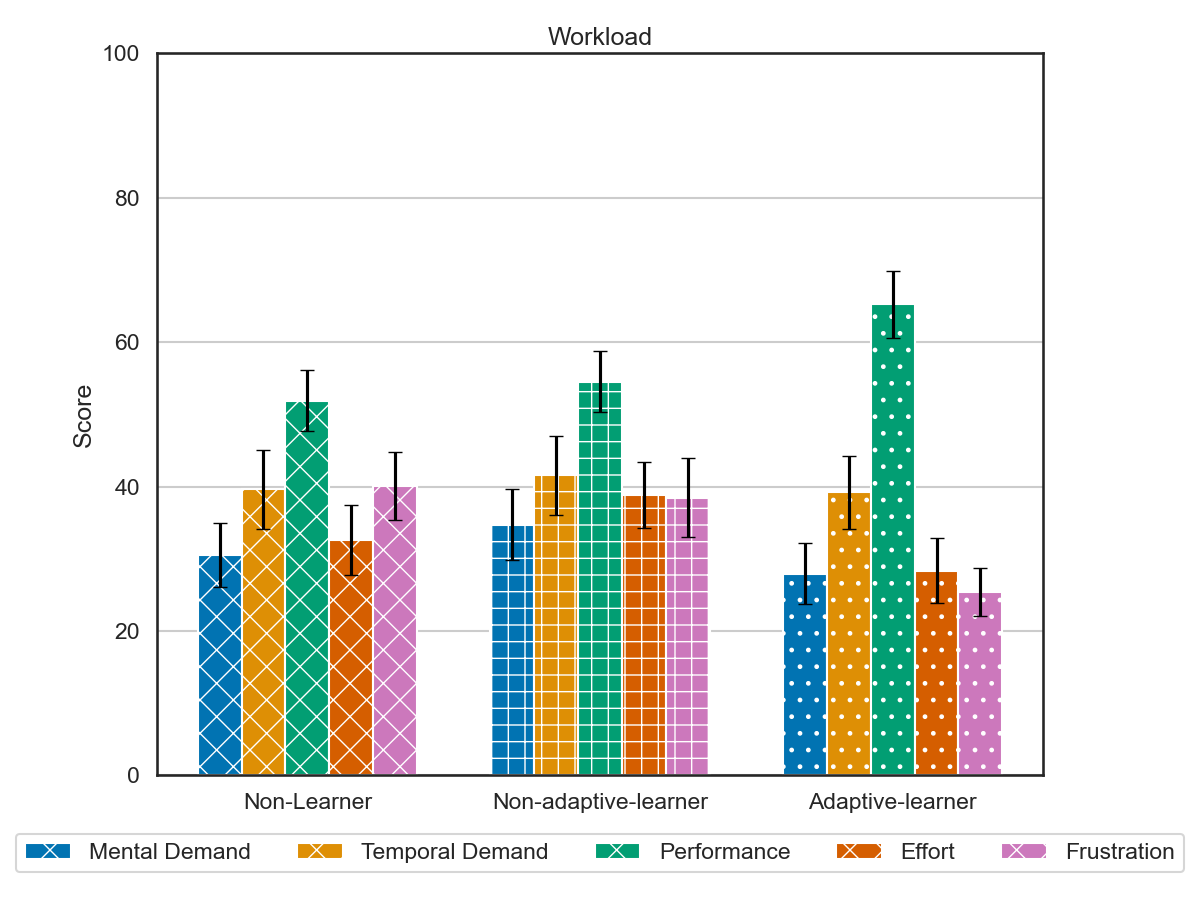}
    \caption{Exp 2: Responses on the NASA TLX scale}
    \label{fig:workload-2b}
\end{figure}

% \subsubsection{Experiment 2:  with uninformed prior.}
Comparing the average workload (\ref{tab:results}) across the three interaction strategies showed a significant difference $(F(2,46)=10872, ~p<0.001)$. 
Figure \ref{fig:workload-2b} shows the participants' responses on each dimension. There were significant differences between the three strategies in performance $(F(2,46)=5.443,~p=0.008)$, effort $(F(2,46)=4.252,~p=0.02)$, and frustration $(F(2,46)=5.454,~p=0.007)$. 
Pairwise comparisons with Bonferroni adjustments showed that the adaptive-learner strategy led to higher perceived performance compared to the non-adaptive learner $(p=0.037)$ and to the non-learner $(p=0.044)$ strategies and led to lower frustration compared to the non-adaptive learner $(p=0.032)$ and to the non-learner $(p=0.017)$ strategies. 

\section{Discussion - Empirical Study}

In this study, we empirically verified the results of the simulation study by having participants interact with non-adaptive strategies which was not value aligned with them and an adaptive strategy that continuously adapts to the human's values during interaction. Results from this study correspond to the case with a high level of risk $d=0.575$ from the simulation study in which, we saw that adapting to human values kept trust high while trust is lost while interacting with a misaligned robot. 

Additionally, we also observed that human workload is lower when interacting with the adaptive-learner strategy compared to that when interacting with the non-learner and non-adaptive learner strategies. Although the objective performance was the same across the three strategies, the participants perceived that they were performing better with the adaptive-learner strategies. The similar objective performance across the three strategies could be attributed to the information provided to the participants, which was usually enough to make their own decisions, regardless of the robot's recommendations. Thus, the human action selection policies were very similar across the three strategies. Only the adaptive-learner strategy was able to learn and match this human policy resulting in a higher number of agreements with the robot's recommendations, which we postulate, led to lower workload and higher trust.

\section{Conclusion}

Most prior work in value alignment in HRI \cite{Hadfield-Menell2016, Fisac2020, Yuan2022, mechergui_goal_2024} makes one key assumption: aligning the robot's values to that of the human's is beneficial. This may be true when we look at the human in isolation; the human's rewards would increase in this case. However, when looking at it from a teaming perspective, we do not know if the human's personal values are the best for handling the task at hand \cite{Milli2017, kwon_when_2020, arnold_value_nodate}. Thus, this assumption needs to be checked in the context of human-robot teams. Our study is one of the first to attempt to study this assumption. Our simulation results indicate that value alignment is beneficial for trust only when there is a high level of risk in the task involved. When the level of risk is low, even value misaligned robots can lead to high trust. Our empirical study shows that value alignment under high risk is indeed beneficial for trust and workload. 

We also explore the idea of incorporating a layer of trust within the value alignment problem. We theorize that since there is a tradeoff between degree of value alignment and trust and since it has been shown that trust is a predictor of behavior in humans \cite{Parasuraman1997, Lee2004}, it is important for the robot to try and find a balance between aligning itself to the human's values and maintaining trust. We hope to see more work done towards this goal in other domains like shared control, social robotics, rehabilitation robotics, etc. 

The results of our study should be seen in light of the following limitations. First, we provide a demonstration in the case when there are only two components in the team's reward function. Therefore, we only need to learn the human's preference for one of the two components and can ascertain their relative preference between the two objectives. Our formulation, however, can readily be extended to the case where there are more than two objectives in the team's reward function, with additional computations required to learn and maintain a distribution over each reward weight. 

Second, our task consists of binary actions: using or not using an armored robot. Judging the performance of the recommendations is fairly easy in this case, since we only need to compare the rewards earned for these two actions. In case more than two actions are available, this assessment becomes more difficult. There could be effects like satisficing \cite{radner1975satisficing} and heuristics which would make defining the performance metric difficult. Thus, although the human trust-behavior model can be readily extended to such a case of multiple actions, extending the trust dynamics model is challenging and is an interesting avenue for future research.

Third, in our scenario, there is an expected skewness among the general population to be more concerned about saving health. It would be interesting to study tasks where the two objectives are more balanced. 

Finally, our scenario, which entails a trade-off between ``saving health" and ``saving time", and the decision to use or not use an armored robot, is informed by the complex decision-making scenarios in real-life HRI contexts, such as DARPA’s SQUAD-X program in which individuals receive recommendations from air and ground robots for various tasks and Shield AI's NOVA 2 program involving the use of small drones for surveillance and decision support. While our scenario offers insights into these types of decisions, we recognize it as a simplified representation of situations where decisions involve numerous objectives, a variety of recommendations, and possible actions. Therefore, further research is essential to determine the applicability of our findings in more complex, real-world environments and to validate the robustness of our conclusions in diverse and dynamic HRI settings.

\begin{acknowledgement}
This work was supported by the Air Force Office of Scientific Research under Grants FA9550-20-1-0406 and FA9550-23-1-0044.
\end{acknowledgement}
%
% \section*{Appendix}
% \addcontentsline{toc}{section}{Appendix}
% %
% %
% When placed at the end of a chapter or contribution (as opposed to at the end of the book), the numbering of tables, figures, and equations in the appendix section continues on from that in the main text. Hence please \textit{do not} use the \verb|appendix| command when writing an appendix at the end of your chapter or contribution. If there is only one the appendix is designated ``Appendix'', or ``Appendix 1'', or ``Appendix 2'', etc. if there is more than one.

% \begin{equation}
% a \times b = c
% \end{equation}

% \input{references}

% \bibliographystyle{spphys.bst}
\bibliographystyle{spbasic.bst}
\bibliography{bibliography}
\end{document}